# Learning Isometric Embeddings of Road Networks using Multidimensional Scaling

Juan Carlos Climent Pardo
*Technical University of Munich*
jc.climent-pardo@tum.de

*Abstract*—The lack of generalization in learning-based autonomous driving applications is shown by the narrow range of road scenarios that vehicles can currently cover. A generalizable approach should capture many distinct road structures and topologies, as well as consider traffic participants, and dynamic changes in the environment, so that vehicles can navigate and perform motion planning tasks even in the most difficult situations. Designing suitable feature spaces for neural network-based motion planers that encapsulate all kinds of road scenarios is still an open research challenge. This paper tackles this learning-based generalization challenge and shows how graph representations of road networks can be leveraged by using multidimensional scaling (MDS) techniques in order to obtain such feature spaces. State-of-the-art graph representations and MDS approaches are analyzed for the autonomous driving use case. Finally, the option of embedding graph nodes is discussed in order to perform easier learning procedures and obtain dimensionality reduction.

## I. Introduction

Advances in computing power through improved hardware, the development of efficient algorithms, as well as access to a greater amount of data are among some of the factors that have driven the automotive sector towards automation of vehicles in recent years [1]–[3]. This, accompanied by the current increase in deep learning techniques (see figure 1) [4], has led to the popularity of autonomous technologies assisting in the transportation field [5]–[8]. Many studies show how learning-based approaches and statistical models, without following explicit instructions, can help vehicles become more autonomous, helping the crucial task of motion planning through traffic scenes, and allowing vehicles to adapt to the environment without harming other traffic participants [9]–[12]. Some might argue that as autonomous technology advances, the idea of a fully autonomous car becomes reality [13], transforming mobility with increased efficiency and personal safety [12], as well as revolutionizing city infrastructures and reducing environmental damage. A "level five" on the autonomy scale defined by the Society of Automotive Engineers (SAE) is foreseen in the early 2030's [14], which corresponds to full automation: being operational anytime, anywhere in the world [15].

However, this also shows that a real-world implementation of fully autonomous vehicles is still in the works and far from finished. One fundamental attribute of these intelligent systems is the ability to learn via training. The abundance of freely available datasets (e.g. [16]–[19]) offers many options to learn, from road networks to possible vehicle trajectories and vehicle interactions, but practical learning-based autonomous driving models must yet become proficient in generalizing behaviors and adjusting to the different scenarios [20]. Most of the previous research done in the learning field is often limited to specific applications and to a narrow range of traffic scenarios. [21]–[24], and [12], for example, only focus on how to perform safe lane-changing maneuvers in highway scenarios. But the need for generalization is a ubiquitous challenge in all sections of autonomous driving [25], [26], from image classification [27] to motion planning [28], as vehicles must have the ability to properly adapt to new, previously unseen data. The moving agent should consider road intersections, road merging, unmarked roads, and other traffic elements when performing the motion planning task. On that note, well-generalizing models will learn from the diversity in the data in order to make good predictions [29], which will allow them to adjust to the different scenarios. This makes navigation in a wider variety of real-world traffic situations possible. Learning-based generalization is often considered as safety-critical in real-world applications [28], [30], and therefore, learning-based robotic applications usually provide fallback mechanisms that guarantee a safe functionality in complicated scenarios [31]–[34], [12]. Because of that, safety verification and validation procedures have become a big research topic nowadays [35], [36], [28].

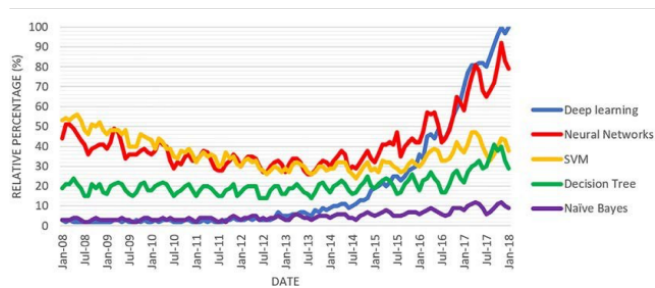

Figure 1: Google trend data showing the growing interest in the machine learning field [4].

The underlying road geometries of traffic scenarios, which influence the spatial relationships between vehicles, as well as the variations in the traffic scenes (dynamic aspects of the locations, participants, etc.) can only be processed by learning-based approaches when represented in a high-level format. This can be done via graph structures [37], as neural networks can practically operate on graph-structured domains [38]. The central challenge is incorporating the structural information of the traffic scenes into the graph, and, as the number of nodes and edges in a graph varies when representing different traffic scenes, invariance to node permutation is desired for the learning procedure. Because of that, graph encodings (also called embeddings) are used, which facilitate the implementation of graph information into learning methods.

Let us illustrate the importance of designing an suitable feature space and graph representation with an example: In a highway driving setting, for instance, it would be an appropriate approach to design a feature space within the cartesian space $\mathbb{R}^2$, considering the distances to neighboring vehicles as Euclidean. Ideally, the travel time distances between any vehicles (or any points on the highway per se), can be represented by the physical distances. This way, a graph can be created with the vehicles being the nodes and the edges representing potential interaction effects. However, this approach will yield a misleading representation of the environment in the general case, since the tangible distances in a road network should not be defined by the Euclidean distance metric as computed in $\mathbb{R}^2$, but rather by the travel distances implied by the road geometry. Hence, the cartesian, naive coordinate system equates a vehicle's interaction potential with spatial closeness, which is only true for a narrow subset of cases.

Figure 2 emphasizes this issue by showing some examples where spatial closeness is not proportional to a possible interaction. 2a and 2b depict two scenes, where two vehicles seem to be close considering the Euclidean distance, yet the probability of their interaction or a collision is low compared to what the naive distance implies, since they are located on oppositely directed lanes (2a), or the lane trajectory sets them further apart (2b). 2c is an additional example that highlights this aspect: two vehicles are close to each other, but in neighboring lanes with opposite directions. Assuming that the vehicles adhere to the traffic rules and do not enter opposite lanes, the interaction potential is rather low, compared to what the Euclidean distance would imply. In a highway scenario, for instance, they could also be separated by a median strip, which would further highlight this issue, since an interaction between two vehicles would only be possible if one crashes into the strip. Hence, it can be assumed that the naive coordinate system may include irrelevant connections when representing a traffic scenario as a graph, which would weaken the model's ability to learn vehicle-to-vehicle interactions. This issue is also observed in [21], whereas the proposed approach can lead to better performance achieving heterogeneous coordinates by considering time-distances as a

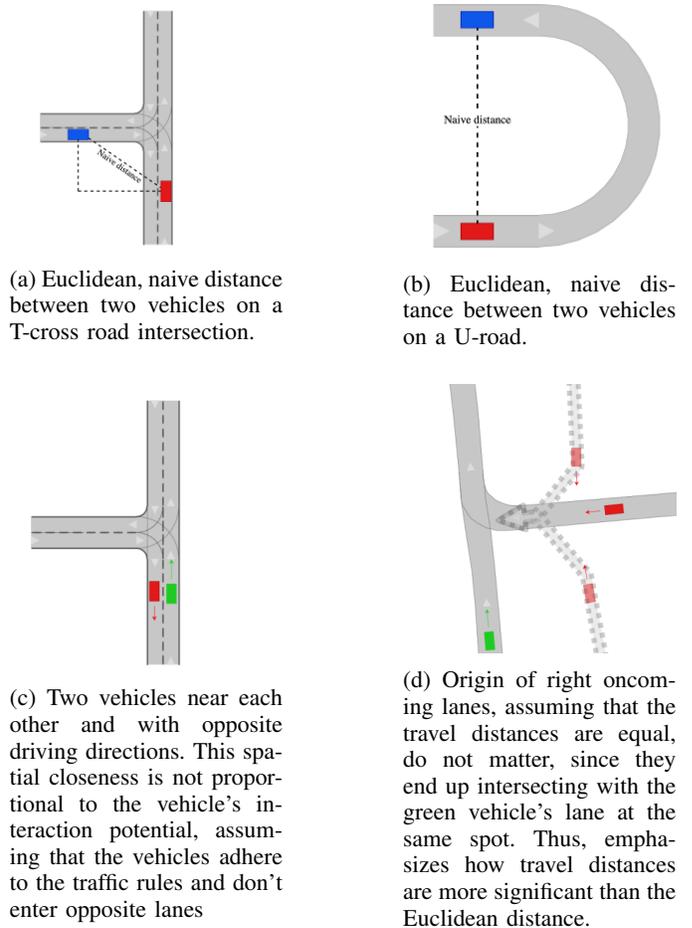

(a) Euclidean, naive distance between two vehicles on a T-cross road intersection.

(b) Euclidean, naive distance between two vehicles on a U-road.

(c) Two vehicles near each other and with opposite driving directions. This spatial closeness is not proportional to the vehicle's interaction potential, assuming that the vehicles adhere to the traffic rules and don't enter opposite lanes

(d) Origin of right oncoming lanes, assuming that the travel distances are equal, do not matter, since they end up intersecting with the green vehicle's lane at the same spot. Thus, emphasizes how travel distances are more significant than the Euclidean distance.

Figure 2: Naive distances and spatial relationships between vehicles in different scenarios.

metric.

*A. Objectives*

In order to solve the optimization of these graphs in a feature space, multidimensional scaling (MDS) techniques are employed. When obtaining the corresponding graph, a transformation of the coordinate system is needed to embed the graph structure and learn it (i.e., as a set of coordinates for each graph node in an embedding space). Generally, MDS receives as an input a set of proximities (e.g., travel distances) for all pairs of some points (or nodes) and attempts to return a configuration of those points in a space of given dimensionality such that the distances between points correspond as closely with the given proximities. Videlicet, MDS generates a matrix of coordinates of a point configuration representing the time-space. Figure 2d helps visualize how MDS handles the travel-distances: It does not matter from the viewpoint of the green ego vehicle where exactly the oncoming lane from the right originates, assuming that the travel distances are equal, since it will end up intersecting with the ego vehicle's lane at the same spot.

Thus, using naive coordinates for defining a feature space for a downstream learning-based autonomous driving algorithm, in this case, would introduce a type of undesired disturbance, since it does not carry any value from a behavior planning standpoint. However, with MDS the three hypothetical incoming lanes would be mapped to the same source node, making a planning model invariant to irrelevant geometric variations.

The optimization problem can be formally expressed by a so called stress function, which MDS tries to minimize:

$$\text{stress}(\boldsymbol{X}) = \sum_{i \neq j = 1, \ldots, N} w_{ij} \left(d_{ij} - \|\boldsymbol{X}_i - \boldsymbol{X}_j\|\right)^2 \quad (1)$$

$$\text{with } w_{ij} = d_{ij}^{-\alpha}$$

$$x^* = \arg\min_x (stress(\boldsymbol{X})) \quad (2)$$

,
Here, $\boldsymbol{X}$ contains the coordinates of each vertex (graph node) in the embedding space (low-dimensional), $d_{ij}$ represents the real distance between them and $w_{ij}$ is a weighting factor used to either emphasize or dampen the importance of certain pairs. According to [39], common practice is to define $d_{ij}$ as the shortest path distance between vertices $i$ and $j$ and $w_{ij} = d_{ij}^{-\alpha}$ to offset the extra weight given to longer path due to squaring the difference. According to Kamada and Kawai, the appropriate value is $\alpha = 2$ [40], but Cohen [41] also considers the possibilities of $\alpha = 0$ or $\alpha = 1$. Cohen also suggested to set $d_{ij}$ equal to the linear-network distance in order to cluster the structures of graphs.

As briefly hinted, the output of MDS is a mapping of node coordinates to an embedded space which preserves distances. The main challenge is to find a suited transformation where the distortion between the actual distances in the underlying road network and the ones in the embedding space is as little as possible. Considering travel distances as better suited to engineer a feature space that encodes road networks into graphs, the output of MDS results in time-distance maps (TDMs), which are mappings based on the travel times on a network. With TDMs, the physical space is deformed so that the edges (e.g., roads) between the nodes (e.g., intersections on the road network) of the graph are proportional to their travel times. In other words, TDMs aim at obtaining a spatial configuration that aligns the distances (Euclidean or not) between consecutive points with the corresponding time-distances [42]. A simple model to calculate time distances would be to consider speed limits as proxies, which would give the relationship between two neighbouring graph nodes (e.g. road elements). As TDMs consider time instead of physical length as a unit of measurement, it has to be noted, that the deforming of space also yields an error, because it is not possible to project time-space onto a two-dimensional plane as accurately as with physical space [43]. This error is normally due to traffic congestion, physical barriers (e.g., road constructions, mountains), or other traffic elements. Last but not least, time-space maps can be represented by isochrones approaches, which are based on time-distances between one central point and many other surrounding ones, and isometric, which also consider the time-distances between the surrounding points.

The primary goal of this paper is to explore the integration options of such graph representations with MDS techniques in order to facilitate downstream prediction and motion planning research. For that, specific multidimensional scaling techniques are reviewed and classified in further sections. Finally, the paper also discusses how graph nodes can be embedded in order to optimize the learning procedure of these trajectory and motion planning tasks. 0

*B. Related Work*

The mentioned previous work in the learning field is often limited to a narrow range of traffic scenarios, as is shown in [21]–[24], and [12], and not focused on generalizability. Exemplary, [21] only considers highway lane-changing scenarios with a varying number of vehicles, and the generalization capabilities of the reinforcement learning algorithm are evaluated under ablation studies. [24] also analyzes multi-lane cruising using a hierarchical planning approach, but again, the scenarios used are restricted to certain types. Furthermore, some of these works hint that something like MDS should be used to advance the state of learning-based autonomous driving. To the best of our knowledge, there is no previous attempt to combine multidimensional scaling techniques with learning approaches in the autonomous driving field (and consider specific embeddings of road networks to learn them) and hence, the motivation of pioneering in this field.

*C. Paper Organization*

The remainder of this paper's structure is organized as follows. In Section II a detailed survey on previous approaches for extracting graph representations of road networks is presented. This can be considered a necessary pre-processing step for using an MDS algorithm afterwards. Section III focuses on the methodology behind MDS techniques and cover different MDS approaches used so far. Furthermore, it introduces how the embeddings can be learned. Finally, Section IV summarizes the paper and concludes the work by discussing and visualizing some application examples.

II. GRAPH REPRESENTATION APPROACHES FOR ROAD NETWORKS

Traffic scenes are composed of three central components: road networks, traffic participants (e.g. vehicles, humans, etc.), and traffic control protocols. Since this paper is an early exploration of the possibilities that graphs and MDS can offer for motion planning tasks, only road networks are considered in this section. Their representation as graphs can be considered as a pre-processing step in the overall embedding learning pipeline. Further expansions to traffic participants are discussed at the end of this work.

As highlighted before, road networks are composed of different complex elements, such as intersections, road merging, unmarked roads, or traffic hazards. To apply learning algorithms to such traffic scenes, the scenarios must be described and interpreted [44]. There are many approaches to this task, as knowledge-based approaches [45], ontology-based approaches with contextual information [46], [47], or tensor-based ones [48]. Nevertheless, graph representations as shown in the work of Ulbrich et al. [49], Diehl et al. [50], or Zipfl and Zoellner [51] are better suited for learning algorithms since they provide a high-level representation while being computationally efficient and providing large model capacities.

*A. Graph Structures & Approaches to Graph Extraction*

First, let us follow graph construction based the subdivision of road networks into smaller segments (i.e. "chunks" of roads). As stated in [37], these blocks facilitate the interpretation of which specific road-parts are being considered at a given point in time. This splitting is also seen in environment benchmarking, like with the composable *CommonRoad* [52] benchmarks, where they are named Lanelets. These blocks define a non-empty and finite set $\mathcal{V}$ of vertices in a graph. Directed edges (e.g. lines) in the graph represent the connections and relationships between two blocks. Thus, if there is no link between two nodes, no direct travel is possible. They are summarized by a set $\mathcal{E}$, where *arc(v,v')* refers to the connection and travel direction between the two nodes. These sets are visualized in figure 3, where the physical road network (right) is abstracted to a graph (left). Furthermore, considering that each block has a specific number of entries and exits, there has to be at least one outgoing edge for each entry and one incoming edge for each exit. Therefore, two further sets are defined: $\mathcal{V}_N$ for the entries and $\mathcal{V}_X$ for the exits. Combining these sets results in a the formal notation of a road network graph $\mathcal{G}$ of the form $\mathcal{G} = \{\mathcal{V}, \mathcal{E}, \mathcal{V}_N, \mathcal{V}_X\}$. With those components, a road path $\mathcal{P}$ is constructible, which can be seen as a sequence of blocks and edges, and which is delimited by a start and endpoint.

In practice, graphs are constructed directly from images of road networks. Previous work in this field tries leveraging overhead imagery to extract road line segments and to obtain the structure of road networks [53]–[57]. With the improvements of DL, many approaches also try to consider neural networks to perform image segmentation and graph estimation [58], [59] with the goal of achieving large-scale lane-level annotations, which are crucial for autonomous driving. The reasoning for having a lane-level view on road networks is found in [60]–[62]. In addition, other works consider ego-vehicle sensor data to obtain the exact geometry of lane boundaries [63], and a few more consider the problem of extracting the precise location of lanes and boundaries [62], [64]. [65] presents a novel approach for lane topography estimation. It automates the process of extracting annotations from inferred vehicle readings and highlights how the underlying graph structure of lanes can be extracted from multimodal, bird's-eye-view image data. This is depicted in figure 4. Inspired by [66], the algorithm of [65] also trains a neural network to estimate the position and orientation of lanes for a given scene.

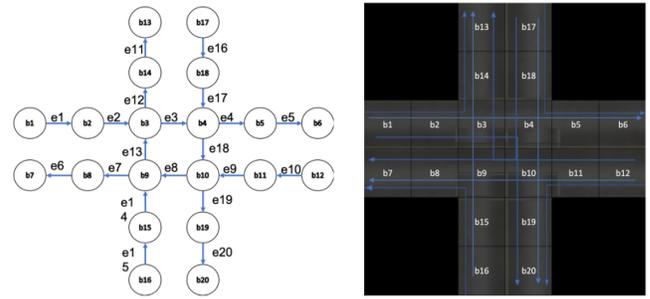

Figure 3: A road graph for four-way intersection [37].

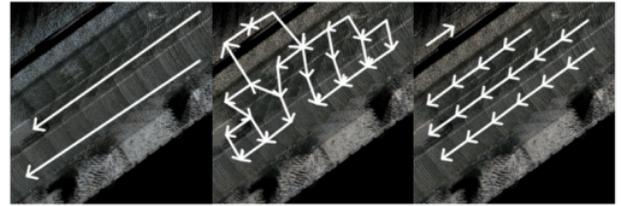

Figure 4: Exemplary predictions of two baseline model B1 (left) and B2 (center). The ground truth graph is visualized on the right [65].

Figure 5 shows an own example of how *CommonRoad Lanelets* are converted to graphs.

### III. METHODOLOGY: MULTIDIMENSIONAL SCALING (MDS)

Once a road network graph representation is obtained, suitable coordinate transformations have to be found, where the distortion between the actual distances in the underlying road network and the ones in the embedding space is as little as possible. This is far from trivial, and through a simple example, the difficulty behind finding a good layout and encoding is illustrated. Consider drawing a tetrahedron on a two-dimensional space, and then try to find an ideal layout where all edges are isometric and have equal lengths. It is not possible with cartesian coordinates. There are too few dimensions available so that sufficient degrees of freedom are provided, even for such a small graph of only four vertices. To get a layout as close as possible, Multidimensional scaling (MDS) is needed. As briefly explained, it is a technique to

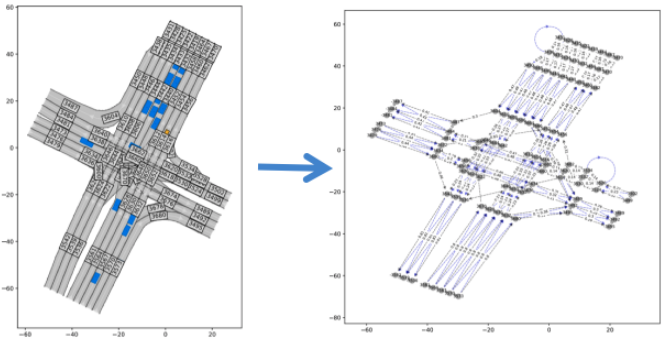

(a) Lanelet intersection considering multiple lanes from [52] (left) is transformed to the corresponding graph structure (right).

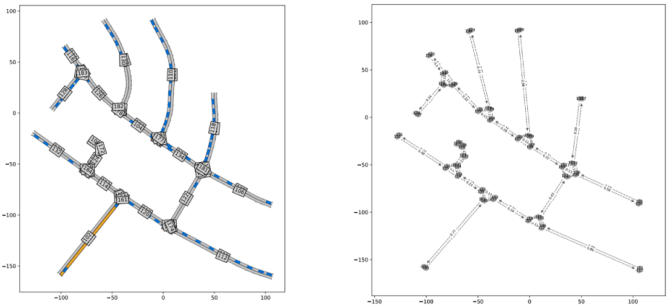

(b) Lanelet layout with multiple intersections from [52] (left) is transformed to the corresponding graph structure (right).

Figure 5

solve this type of problems, which attempts to minimize disparities between ideal and low-dimensional distances. In our learning-based autonomous driving approach, MDS is used on the graphs (and nodes embeddings, which help reducing dimensionalities in the learning process).

### A. MDS Origins & Applications

Originally, MDS techniques were developed in the 70´s and 80´s [67] for psychology studies that dealt with problems of measuring and predicting human judgments. They aligned spatial representations of psychological stimuli with people's judgments (e.g. preferences, assuming similarities, relatedness, etc.) [68], and investigated possible impacts in social [69], clinical [70], and developmental [71] psychology. Nowadays, the applicability has grown across many disciplines such as neuroscience [72], marketing [73], ecology [74], and political sciences [75].

### B. MDS Characteristics

The key aspect of MDS techniques is that no prior knowledge about underlying factors in the graphs corresponding to specific dimensions is needed. Therefore, assumptions as normality and linearity in the data can be ignored [76]. The only needed data are the proximities in distance pairs, which equals to the input being a node distance matrix with these pair-wise proximities. For a graph composed of N items, $(N * (N − 1))/2$ proximities must be acquired, so that each node is compared to every other at least once. This clearly shows how the number of comparisons rapidly grows. Following equation 1 they are then transformed to measures of dissimilarity. The stress function quantifies the amount of conflict that is present in the data and differs somewhat from equation 1 depending on the scaling algorithm used. These dissimilarities can be considered related to the Euclidean distances in some space of unknown dimensions as they measure the agreement between the input proximities and estimated distances. Thus, the more similar the two nodes, the closer they appear in the multidimensional space, and the lower the stress value. Analogously, the more dissimilar, the farther apart both nodes will be. As all distances to all other nodes are given, MDS attempts to fit all dyadic data in a certain least-squares sense, increasing iteratively the fidelity to the input data by minimizing this function. Until a very good fit is found (e.g., loss function shows very little improvement, or optimization criterion is reached), the dimensions can be increased and adjusted [68], [67]. This increase has mainly two effects, it adds a degree of freedom to the movement of individual items and it decreases the stress value of that dimension. The goal is to strike a balance between a good solution of the stress function and an interpretable solution [77]. The conventional dimension for this is considerably below the N-1 dimensions that would be needed to have a perfect fit among N given nodes (as defined in [78]), and the termination criterion is usually a predetermined stress value or a specific number of iterations. In the end, the output of MDS is a mapping of node coordinates to an embedded space, which spatially conveys the relationships among nodes, and which allows inferring the underlying dimensions of the data. Moreover, by assessing the distances on the map, one can obtain a quantitative measurement of their perceived similarity, relative to other nodes in the space. These types of maps do not have an orientation, which implies that a rotation can be performed around the center, as the only important characteristic are the relative positions of each node. Although there is no guarantee that a meaningful set of dimensions will emerge with MDS, the results typically are useful, since the graphical representation facilitates the comprehension of patterns in the data. The relationships are normally visualized in a clear, concise, and informative manner [68].

At a gross level, it is usually differentiated between metric and non-metric multidimensional scaling algorithms, according to the application field, the data collection procedure, and the definition of the stress function. It is a metric one if the scaling function is assumed to be linear when mapping proximities to disparities (which is the case for this autonomous driving applications), and non-metric if it is assumed to be merely (positive) monotonic (or rank order-preserving) [77]. Furthermore, the functions used in MDS are assumed to be increasing or decreasing (whether it is a metric or non-metric approach) depending on if the data are proximities (e.g., similarities) or antiproximities (e.g., dissimilarities). These distances are usually considered Euclidean, but as mentioned before, time distanced are better suited for our autonomous driving applications. Finally there is also a distinction between MDS methods that consider the distances among objects to be symmetric, and the ones where this consideration is not always satisfied, and define similarities as asymmetric. In road networks this can be interpreted for example by the direction of traffic, distances that move forward can be modelled to be positive and distances that move backwards to be negative. Other asymmetric relations are found in cognitive relations [79], [80], where similarities among objects are not always represented by a function of only inter-point distances but also a function of the quantities associated with these objects [81]–[84]. A further extended discussion of model selection is beyond the scope of this brief article, so the reader is advised to consult [85], [86], and the work of Kruskal et al. [87], [88] for further information.

*C. Classical MDS vs. Gradient-Based*

The two main categories of MDS optimization are *classical MDS (CMDS)* approaches and *gradient-based* ones. CMDS was independently proposed in [89], [90] and can be regarded as a spectral graph optimization algorithm. Recent approximation techniques, like PivotMDS [91] make it scale to very large graphs [92]–[94], although spectral methods, in general, have a limited ability to meet specific application requirements. They have a tendency to result in excessive occlusion for certain classes of sparse graphs, and because of that, it is normally recommended to use CMDS as a reliable initialization for distance scaling [95]. CMDS is applied regarding the shortest-path distances $\delta_{ij} = d_G(i,j)$, $i,j \in V$, of undirected graphs $\mathcal{G} = \{\mathcal{V}, \mathcal{E}\}$ [96]. This corresponds to straight-line drawings in D-dimensional spaces, which represent close vertices near to each other and structurally distant ones far apart. The shortest-path CMDS techniques are good at displaying symmetries, but preferred variants focus on minimizing the stress function (equation 1) by exemplary fitting weighted distances directly [91]. It has the advantage that local details are prioritized, as stress minimization is sensitive to local minima [95], [97], and makes the bias towards large distances ideal for initialization purposes. [96] presents a new approach by combining CMDS with PCA, where first the input, rather than the drawing, is adapted, so that more detail is preserved in the periphery, and then, by increasing the number of output dimensions, the coordinates that represent distances more accurately are determined. Finally weighted principal component analysis is used to project the distances into a two- or three-dimensional space, such that local details are favored.

*CMDS* typically uses only a single matrix of disatnces and is not specially considered robust. On the other hand, *replicated MDS* makes use of multiple matrices (matrix decompostion) and is thus more robust thanks to the increased data. *Weighted MDS* is specially used when differences between the individual analyzed items are interesting, as the model treats data from multiple sources, and indicates the degree to which each item weighted the dimensions that are revealed in the common space. An example, therefore, is found in [98], where the most accurate range measurements in sensor networks are weighted for distributed node localization. Although some other terms may be included under the umbrella of the "classical multidimensional scaling", such as *multidimensional unfolding* (similar to weighted MDS but assuming that each item has a vantage point), or *correspondence analysis* (representation of categorical data a the Euclidean space), this should not be a list of MDS models, but show how the models can vary according to application, providing more than just a data visualization with dimensionality reduction.

One of the most used MDS techniques on graphs developed by Gansner et al. [99] is so-called stress *majorization*, which offers a guaranteed monotonic decrease of the energy value, improved robustness against local minima, and shorter running times. As described in the literature of De Leeuw [100] or Groenen et al. [101], and [99], majorization minimizes the function by iteratively analyzing a series of simpler functions and finding their corresponding minima. Each minimum of these simpler functions touches the original function, which serves as an upper bound for them. In other words, equation 1 is decomposed into the individual terms as follows:

$$\text{stress}(X) = \sum_{i<j} w_{ij} \left( \|X_i - X_j\| - d_{ij} \right)^2$$
$$= \sum_{i<j} w_{ij} d_{ij}^2 + \sum_{i<j} w_{ij} \|X_i - X_j\|^2$$
$$- 2 \sum_{i<j} \delta_{ij} \|X_i - X_j\|$$

The first term is a constant independent of the current layout. The second one is a quadratic sum, which can be written using a weighted Laplacian $L^w$, and the third one is bounded via a Cauchy-Schwartz inequality. All terms can then be bounded by a stress function $F^Z(X)$ resulting in:

$$\sum_{i<j} w_{ij} \|X_i - X_j\|^2 = \text{Tr}\left( X^T L^w X \right)$$

$$\sum_{i<j} \delta_{ij} \|X_i - X_j\| \geqslant \text{Tr}\left( X^T L^Z Z \right)$$

$$F^Z(X) = \sum_{i<j} w_{ij} d_{ij}^2 + \text{Tr}\left( X^T L^w X \right) - 2 \text{Tr}\left( X^T L^Z Z \right)$$

$$\text{stress}(X) \leqslant F^Z(X)$$

For a further derivation consult [99]. A practical implementation of majorization in R is seen in [102]. There are three extensions to majorization stress optimization that use its power and flexibility and have established the method as state-of-the-art in the past decade. The first one weights edge lengths so that the drawing area is better used, which is especially useful in real-life graphs with degrees following a power-law degrees distribution. The second one handles sparse stress functions, only considering a small fraction of all pairwise distances, which reduces the time and space complexity of stress optimization. Sparse stress optimization is practically impossible when using the Kamada-Kawai [40] technique unless a very good initialization is available. Finally, the third extension obtains an approximation of the graph by setting constraints on the axes of the vector space (or rather a subspace) it is supposed to be constructed (see [103] for more information).

The second group of MDS methods is the *gradient-based* approaches, which were originated by Kruskal [104]. In [105], a stochastic gradient descent (SGD) algorithm is proposed, which approximates the gradient of the stress function via a sum of the gradients of its terms and thus updates a single pair of vertices at a iteration. It reaches lower stress levels faster and more consistently than *majorization* and does not need a good initialization. Furthermore, it makes it easier to produce constrained layouts than previous approaches and also includes the possibility of making SGD scalable to large graphs by handling the sparse stress approximation of Ortmann et al. [106]. The main challenges when working with SGD are selecting an appropriate step size and the number of iterations if time is a limiting factor (consider a limited subset of possible schedules).

In order compare both methods, [105] follows Khoury et al. [107] by using symmetric sparse matrices from the SuiteSparse Matrix Collection [108] as a benchmark, and obtains the following plots:

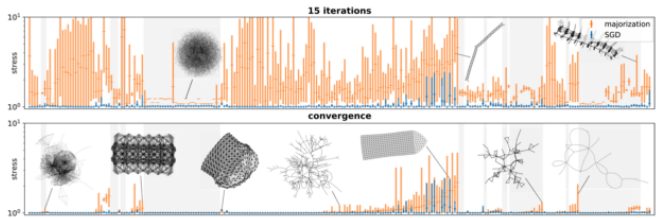

Figure 6: Stress values after applying SDG on the SparseSuite Matrix Collection [108] for 15 iterations (top) and after convergence (bottom). The shown markers indicate the mean over all runs, with bars ranging from minimum to maximum on any run [105]. The models are representative of different graph types, with varying in mean degree, and structure forms (with twists, densely packed, etc.).

Figure 6 shows how SGD reaches almost equal low-stress levels on almost every run. Although majorization is proven to monotonically decrease stress [99], it can sometimes struggle with local minima, which is depicted by figure 7. Here, majorization consistently shows the larger variance in its stress trajectories from different starting configurations. Finally, figure 8 depicts how SDG converges to low-stress levels in far fewer iterations than majorization. The most global minima are found after a few iterations, which makes the algorithm attractive for real-time applications.

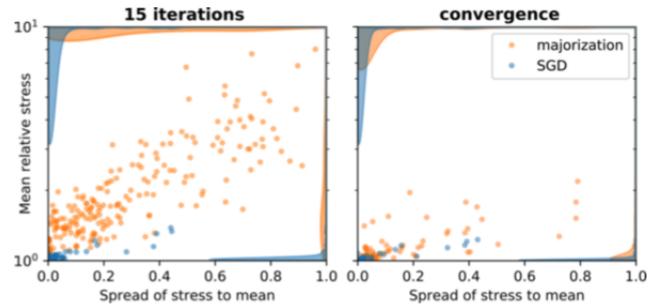

Figure 7: Both scatter plots show the mean stress relative to the best achieved, against the spread of the stress. This is measured as the standard deviation over mean (= coefficient of variation) and the plots show the results after 15 iterations (left) and after convergence (right) [105].

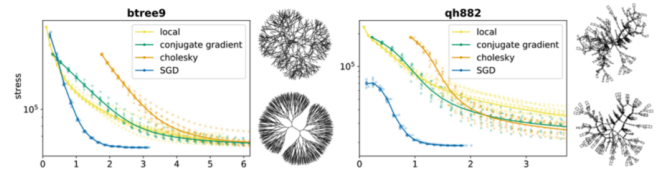

Figure 8: Speed comparison of different optimization with two graph models of the mentioned dataset. Note that the Cholesky factorization from [109] requires a long initialization which is visible through he horizontal offset of the curves.

### D. Graph Encoding and Learning

The previous section showed the ubiquity of graphs, as they are key features in countless types of systems, efficiently storing and enabling access to relational data [110]. Once the corresponding graph is generated from the scene readings, it can be embedded, optimized, and learned. The main idea is to learn a mapping that embeds nodes or subgraphs into a low-dimensional vector space $\mathbb{R}^d$, and then optimize this mapping via MDS so that geometric relationships in this learned latent space reflect the structure of the original graph. Then, these learned embeddings can be used as input features for downstream prediction or motion planning tasks. The central problem is to incorporate the graph structure information into a machine learning model. As described in [111], there have to be considered two types of graph learning approaches: supervised and unsupervised learning. In supervised representation learning, the models use classification and regression labels (associated with individual nodes or entire subgraphs) to optimize the embeddings of graph nodes. In an unsupervised manner, without knowing anything about the further machine learning tasks, the mapping of the embeddings is learned.

The discussion about the learning categories begins with the node embedding itself. The main goal is to summarize the graph position of the nodes and the structures of a node's local graph neighborhood by encoding them as low-dimensional vectors. The projection of nodes into a latent space implies a correlation between the geometric relations in this low-dimensional space and the interactions (edges) in the original graph [112]. Vectornet [113] and LaneGCN [114] propose approaches that encode the roads using high-order vector representations, graph attention networks and graph convolutions to efficiently replace conventional rendering-based models. According to [111], to methodologically classify the different approaches and put them in the same notation, first, an encoding and decoding framework has to be organized.

The encoder maps each node to a low-dimensional vector, and the decoder obtains the structural information about the graph from those embeddings. If the embeddings are done properly and contain all necessary information for downstream machine learning tasks, then there is a way to decode the high-dimensional graph information—such as the global positions of nodes in the graph and the structure of local graph neighborhoods. As an example, the decoder might predict, given the embeddings, the existence of edges between nodes. Normally, a pairwise decoder is used [115], mapping pairs of node embeddings to real-valued graph measures, and quantifying the proximity of nodes in the original graph. Afterwards, to optimize the encoder-decoder system, a loss function determining the quality of the pairwise reconstructions is evaluated, and the model trained. Most of the cases use stochastic gradient descent [105] for optimization, though some algorithms do permit closed-form solutions via matrix decomposition [116]. Once the encoder-decoder system is optimized, the trained encoder can generate embeddings for nodes, which then can be used as feature inputs for the motion planning tasks.

Most of the node embeddings rely on so-called direct encodings. An overview of well-known direct encoding algorithms is presented in [111]. They can be classified into matrix factorization methods such as Laplacian Eigenmaps [117], GraRep [116] or Graph Factorization [115], and random walk approaches such as the DeepWalk model of Perozzi et al. [118] or node2vec [119]. The general encoder function maps nodes to vector embeddings via an "embedding lookup" of the form:

$$\text{ENC}(v_i) = \mathbf{Z}\mathbf{v}_i \quad (3)$$

where $\mathbf{Z} \in \mathbb{R}^{d \times |\mathcal{V}|}$ is a matrix containing the embedding vectors for all nodes and $v_i \in \mathbb{I}_\mathcal{V}$ is a one-hot indicator vector indicating the column of $\mathbf{Z}$ corresponding to node $v_i$.

Since these direct encodings have some drawbacks, such as parameters not being shared between nodes in the encoder or being inherently transductive [120], i.e., they can only generate embeddings for nodes present while training, more complex encoding approaches inside of the encoder-decoder framework exist. These are neighbourhood autoencoder methods as presented in [121]–[123], and neighbourhood aggregation and convolutional encoders [124], [125]. Both, direct encodings and the more complex auto-encoder structures are motivated by matrix factorization techniques and dimensionality reduction. Because of that, MDS is suited to make use of the embeddings when optimizing the graph. On that optimization note, [126] presents *Glimmer*, a multilevel algorithm for MDS designed to exploit GPU features and improve the speed of computation as well as the visual quality of representations. The algorithm organizes input into a hierarchy of levels, which makes the algorithms getting stuck in local minima less likely, and then recursively applies GPU techniques to perform computation parallelisms. The benchmarking of *Glimmer* is done against some of the already mentioned techniques like *SMACOF*, *PivotMDS*, or *CMDS*.

### E. Notable MDS applications

This subsection presents two interesting MDS projects, which have had a lot of relevance in the last years. [127] is another MDS-based localization approach, similar to [98], which considers three different categories of MDS (centralized, semi-centralized, and distributed MDS) for range-free localization, only considering proximity information of sensors for wireless networks. The three types of MDS vary in how they handle computational complexity and localization error optimization. On the other hand, [128] presents an approach on how to perform traffic flow prediction in urban networks by adjusting MDS in three steps. First, the data selection is based on quantitive analysis, second, the data is grouped via MDS and third dimensionality reduction is done based on the correlation coefficient of nodes. Therefore, multiple linear regression models and backpropagation neural networks are employed.

## IV. DISCUSSION & CONCLUSION

An approach was presented that leverages graph representations of road networks by combining them with MDS techniques that take pair-wise proximities of graph nodes and optimizes the representation by using node embeddings. The main challenges in this project are the selection of an appropriate embedding type after having obtained a graph representation, as well as the control of the embedding speed and accuracy. When building a learning pipeline, one should always consider possible bottlenecks in the procedure, as those are especially interesting when performing an online algorithm. Since the main focus of this research is generalizability, both for variation in the traffic scene, as well as for the underlying road geometry, a large number of unique road segments during the model training are considered. For that reason, the speed at which the coordinate embeddings can be computed or trained is quite a significant factor. As accuracy is also important to achieve plausible results, a trade-off between both parameters has to be pondered. This balancing of both factors is a critical feature when applying it to diverse scenarios and because of that, different optimization approaches had to be evaluated. The most prominent was *gradient-based* MDS that showed faster convergence and better stress values after a few iterations.

Finally visualization is provided with figure 9 and figure 10 on how MDS algorithms can be applied to road network graph embeddings, by using a non-euclidean target space, that considers time distances instead of physical ones as a metric. It has to be noted, that the optimization is directly done in a projection space following Riemannian Adam [129] with the help of the Pytorch extension library *geoopt* [130]. Furthermore, not only the embeddings but also the curvature $\kappa$ of the $\kappa$-stereographic embedding space is learnable and a hyperparameter, which gives the representations more flexibility. A thorough explanation on non-euclidean spaces is found in [131].

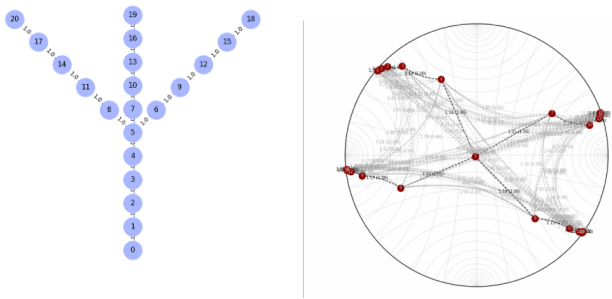

Figure 9: Tree graph (left) optimized with PyTorch extension library for manifold aware gradient descent *geoopt* [130]. Result (right) shows that nodes that seemed close, for example 15 to 16 in the tree graph, are actually far away in the hyperbolic space.

A further expansion of this work would be to consider vehicles and traffic protocols in addition to the road networks. Vehicles, for example, could be mapped into the learned embedded space by defining projection functions on the edges of the graphs. This way, they could be assigned to road entries and exits, paths, and to time points where a vehicle is expected to enter a road segment. Traffic protocols would have to defined so that the general traffic rules can be preserved, and could be classified in time-based traffic controls (which specify which road segments are occupied or free at each time point) and priority-based ones (which specify if constraint arcs are compatible or not). These aspects could drive the presented approach towards a more universal behavior.

## References

[1] Ekim Yurtsever, Jacob Lambert, Alexander Carballo, and Kazuya Takeda. A survey of autonomous driving: Common practices and emerging technologies. *CoRR*, abs/1906.05113, 2019.

[2] C. Thorpe, M. Herbert, T. Kanade, and S. Shafer. Toward autonomous driving: the cmu navlab. i. perception. *IEEE Expert*, 6(4):31–42, 1991.

[3] C. Thorpe, M. Herbert, T. Kanade, and S. Shafter. Toward autonomous driving: the cmu navlab. ii. architecture and systems. *IEEE Expert*, 6(4):44–52, 1991.

[4] ResearchGate. Google trends data for machine learning algorithms between 2008 and 2018, Last Access: 04.02.2022. https://www.researchgate.net/figure/Google-trends-data-for-machine-learning-algorithms-between-2008-and-2018_fig1_334435214.

[5] T. Banerjee, S. Bose, A. Chakraborty, T. Samadder, Bhaskar Kumar, and T. K. Rana. Self driving cars: A peep into the future. In *2017 8th Annual Industrial Automation and Electromechanical Engineering Conference (IEMECON)*, pages 33–38, 2017.

[6] Victor Talpaert, Ibrahim Sobh, B. Ravi Kiran, Patrick Mannion, Senthil Kumar Yogamani, Ahmad El Sallab, and Patrick Pérez. Exploring applications of deep reinforcement learning for real-world autonomous driving systems. *CoRR*, abs/1901.01536, 2019.

[7] Bangalore Ravi Kiran, Ibrahim Sobh, Victor Talpaert, Patrick Mannion, Ahmad A. Al Sallab, Senthil Kumar Yogamani, and Patrick Pérez. Deep reinforcement learning for autonomous driving: A survey. *CoRR*, abs/2002.00444, 2020.

[8] Fei Ye, Shen Zhang, Pin Wang, and Ching-Yao Chan. A survey of deep reinforcement learning algorithms for motion planning and control of autonomous vehicles. *CoRR*, abs/2105.14218, 2021.

[9] David González, Joshué Pérez, Vicente Milanés, and Fawzi Nashashibi. A review of motion planning techniques for automated vehicles. *IEEE Transactions on Intelligent Transportation Systems*, 17(4):1135–1145, 2016.

[10] Brian Paden, Michal Čáp, Sze Zheng Yong, Dmitry Yershov, and Emilio Frazzoli. A survey of motion planning and control techniques for self-driving urban vehicles. *IEEE Transactions on Intelligent Vehicles*, 1(1):33–55, 2016.

[11] Wilko Schwarting, Javier Alonso-Mora, and Daniela Rus. Planning and decision-making for autonomous vehicles. *Annual Review of Control, Robotics, and Autonomous Systems*, 1(1):187–210, 2018.

[12] Hanna Krasowski, Xiao Wang, and Matthias Althoff. Safe reinforcement learning for autonomous lane changing using set-based prediction. In *2020 IEEE 23rd International Conference on Intelligent Transportation Systems (ITSC)*, pages 1–7, 2020.

[13] Murat Dikmen and Catherine Burns. Autonomous driving in the real world: Experiences with tesla autopilot and summon. pages 225–228, 10 2016.

[14] McKinsey Center for Future Mobility. We help leaders across all sectors relevant to the autonomous-driving ecosystem develop a deeper understanding of the disruptions and opportunities ahead., accessed: 30.12.2021.

[15] Philip E. Ross. Robot, you can drive my car. *IEEE Spectrum*, 51(6):60–90, 2014.

[16] Julian Bock, Robert Krajewski, Tobias Moers, Steffen Runde, Lennart Vater, and Lutz Eckstein. The ind dataset: A drone dataset of naturalistic road user trajectories at german intersections. 2019.

[17] Robert Krajewski, Julian Bock, Laurent Kloeker, and Lutz Eckstein. The highd dataset: A drone dataset of naturalistic vehicle trajectories on german highways for validation of highly automated driving systems. In *2018 21st International Conference on Intelligent Transportation Systems (ITSC)*, pages 2118–2125, 2018.

[18] Wei Zhan, Liting Sun, Di Wang, Haojie Shi, Aubrey Clausse, Maximilian Naumann, Julius Kümmerle, Hendrik Königshof, Christoph Stiller, Arnaud de La Fortelle, and Masayoshi Tomizuka. INTERACTION Dataset: An INTERnational, Adversarial and Cooperative moTION Dataset in Interactive Driving Scenarios with Semantic Maps. *arXiv:1910.03088 [cs, eess]*, 2019.

[19] Pei Sun, Henrik Kretzschmar, Xerxes Dotiwalla, Aurelien Chouard, Vijaysai Patnaik, Paul Tsui, James Guo, Yin Zhou, Yuning Chai, Benjamin Caine, et al. Scalability in perception for autonomous driving: Waymo open dataset. In *Proceedings of the IEEE/CVF Conference on Computer Vision and Pattern Recognition*, pages 2446–2454, 2020.

[20] Shivam Akhauri, Laura Zheng, Tom Goldstein, and Ming Lin. Improving generalization of transfer learning across domains using spatio-temporal features in autonomous driving, 2021.

[21] Patrick Hart and Alois C. Knoll. Graph neural networks and reinforcement learning for behavior generation in semantic environments. *CoRR*, abs/2006.12576, 2020.

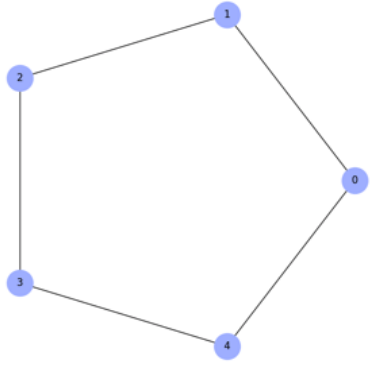

(a) Original graph (pentagon).

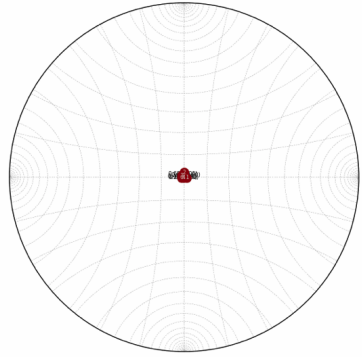

(b) $\kappa = -0.005, loss = 0.332$

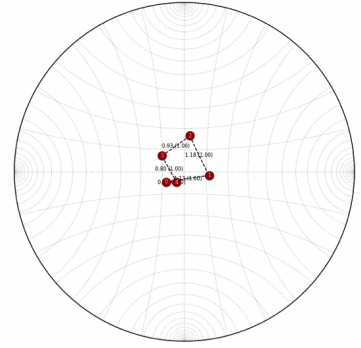

(c) $\kappa = -0.205, loss = 0.220$

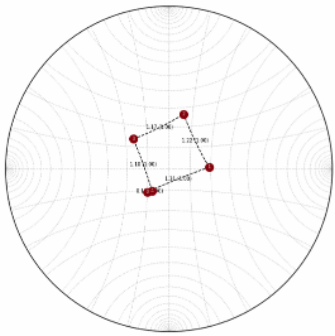

(d) $\kappa = -0.404, loss = 0.140$

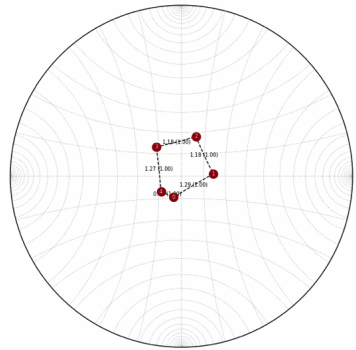

(e) $\kappa = -0.181, loss = 0.081$

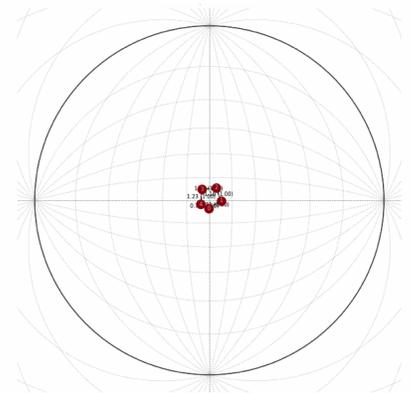

(f) $\kappa = 0.019, loss = 0.040$

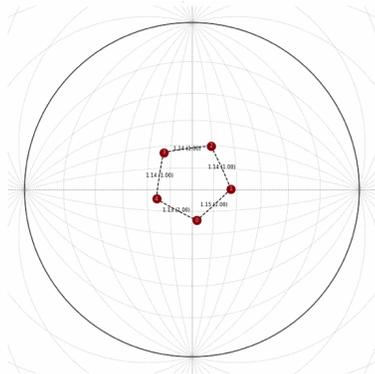

(g) $\kappa = 0.219, loss = 0.017$

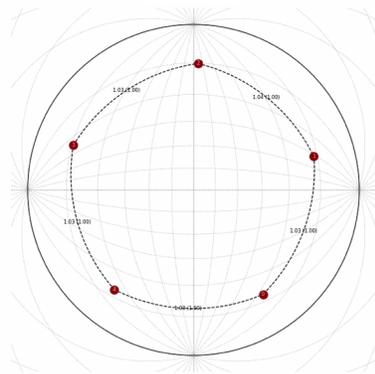

(h) $\kappa = 1.367, loss = 0.001$

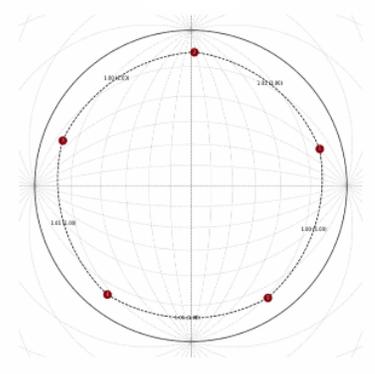

(i) $\kappa = 1.517, loss = 0.001$

Figure 10: Selected steps of the iterative optimization process of a pentagon graph with the PyTorch extension library for manifold aware gradient descent optimization *geoopt* [130]. Each node in the graph is represented by a two-dimensional coordinate vector in a $\kappa$-stereographic embedding space. At each step (b)–(i) the curvature $\kappa$ and the loss are optimized in a joint fashion. In the end, the pentagon assumes a spherical form because of the hyperbolical space it is being optimized on.

[22] Maria Hügle, Gabriel Kalweit, Moritz Werling, and Joschka Boedecker. Deep surrogate q-learning for autonomous driving. *CoRR*, abs/2010.11278, 2020.

[23] Maria Hügle, Gabriel Kalweit, Moritz Werling, and Joschka Boedecker. Dynamic interaction-aware scene understanding for reinforcement learning in autonomous driving. *CoRR*, abs/1909.13582, 2019.

[24] Kasra Rezaee, Peyman Yadmellat, Masoud S. Nosrati, Elmira Amirloo Abolfathi, Mohammed Elmahgiubi, and Jun Luo. Multi-lane cruising using hierarchical planning and reinforcement learning. *CoRR*, abs/2110.00650, 2021.

[25] Mohamed Elsayed, Kimia Hassanzadeh, Nhat M Nguyen, Montgomery Alban, Xiru Zhu, Daniel Graves, and Jun Luo. Ultra: A reinforcement

[26] Isaac Han, Dong-Hyeok Park, and Kyung-Joong Kim. A new open-source off-road environment for benchmark generalization of autonomous driving. *IEEE Access*, 9:136071–136082, 2021.
[27] Hazem Rashed, Eslam Mohamed, Ganesh Sistu, Varun Ravi Kumar, Ciaran Eising, Ahmad El-Sallab, and Senthil Yogamani. Generalized object detection on fisheye cameras for autonomous driving: Dataset, representations and baseline. In *Proceedings of the IEEE/CVF Winter Conference on Applications of Computer Vision*, pages 2272–2280, 2021.
[28] Jaime F. Fisac, Anayo K. Akametalu, Melanie Nicole Zeilinger, Shahab Kaynama, Jeremy H. Gillula, and Claire J. Tomlin. A general safety framework for learning-based control in uncertain robotic systems. *CoRR*, abs/1705.01292, 2017.
[29] Emil Praun and Michael Benisch. Generalization in autonomous vehicle development, accessed: 31.12.2021.
[30] Sandy H. Huang, Nicolas Papernot, Ian J. Goodfellow, Yan Duan, and Pieter Abbeel. Adversarial attacks on neural network policies. *CoRR*, abs/1702.02284, 2017.
[31] P. Abbeel, Adam Coates, Morgan Quigley, and A. Ng. An application of reinforcement learning to aerobatic helicopter flight. In *NIPS*, 2006.
[32] Adam Coates, Pieter Abbeel, and Andrew Y. Ng. Learning for control from multiple demonstrations. In *Proceedings of the 25th International Conference on Machine Learning*, ICML '08, page 144–151, New York, NY, USA, 2008. Association for Computing Machinery.
[33] J. Zico Kolter, Christian Plagemann, David T. Jackson, Andrew Y. Ng, and Sebastian Thrun. A probabilistic approach to mixed open-loop and closed-loop control, with application to extreme autonomous driving. In *2010 IEEE International Conference on Robotics and Automation*, pages 839–845, 2010.
[34] Sergei Lupashin, Angela Schoellig, Michael Sherback, and Raffaello D'Andrea. A simple learning strategy for high-speed quadrocopter multi-flips. In *2010 IEEE International Conference on Robotics and Automation*, pages 1642–1648, 2010.
[35] Walther Wachenfeld and Hermann Winner. *The Release of Autonomous Vehicles*, pages 425–449. 05 2016.
[36] Zhiqi Feng, Wenjie Song, Mengyin Fu, Yi Yang, and Meiling Wang. Decision-making and path planning for highway autonomous driving based on spatio-temporal lane-change gaps. *IEEE Systems Journal*, pages 1–11, 2021.
[37] Jianglin Qiao, Dongmo Zhang, and Dave Jonge. *Graph Representation of Road and Traffic for Autonomous Driving*, pages 377–384. 08 2019.
[38] Jie Zhou, Ganqu Cui, Zhengyan Zhang, Cheng Yang, Zhiyuan Liu, and Maosong Sun. Graph neural networks: A review of methods and applications. *CoRR*, abs/1812.08434, 2018.
[39] Ulrik Brandes and Christian Pich. An experimental study on distance-based graph drawing. 09 2008.
[40] Tomihisa Kamada, Satoru Kawai, et al. An algorithm for drawing general undirected graphs. *Information processing letters*, 31(1):7–15, 1989.
[41] Jonathan D Cohen. Drawing graphs to convey proximity: An incremental arrangement method. *ACM Transactions on Computer-Human Interaction (TOCHI)*, 4(3):197–229, 1997.
[42] Eihan Shimizu and Ryo Inoue. Time-distance mapping: visualization of transportation level of service. In *In Proc. of Symposium on Environmental Issues Related to Infrastructure Development*, pages 221–230, 2003. 00014.
[43] Eihan Shimizu. TIME-SPACE MAPPING BASED ON TOPOLOGICAL TRANSFORMATION OF PHYSICAL MAP. page 12. 00029.
[44] Till Menzel, Gerrit Bagschik, and Markus Maurer. Scenarios for development, test and validation of automated vehicles. *CoRR*, abs/1801.08598, 2018.
[45] Daniel Bogdoll, Jasmin Breitenstein, Florian Heidecker, Maarten Bieshaar, Bernhard Sick, Tim Fingscheidt, and J. Marius Zöllner. Description of corner cases in automated driving: Goals and challenges. *CoRR*, abs/2109.09607, 2021.
[46] Gerrit Bagschik, Till Menzel, and Markus Maurer. Ontology based scene creation for the development of automated vehicles. *CoRR*, abs/1704.01006, 2017.
[47] Alexandre Armand, David Filliat, and Javier Ibañez-Guzman. Ontology-based context awareness for driving assistance systems. In *2014 IEEE Intelligent Vehicles Symposium Proceedings*, pages 227–233, 2014.
[48] Dominik Petrich, Darius Azarfar, Florian Kuhnt, and J. Marius Zöllner. The fingerprint of a traffic situation: A semantic relationship tensor for situation description and awareness. In *2018 21st International Conference on Intelligent Transportation Systems (ITSC)*, pages 429–435, 2018.
[49] Simon Ulbrich, Tobias Nothdurft, Markus Maurer, and Peter Hecker. Graph-based context representation, environment modeling and information aggregation for automated driving. In *2014 IEEE Intelligent Vehicles Symposium Proceedings*, pages 541–547, 2014.
[50] Frederik Diehl, Thomas Brunner, Michael Truong-Le, and Alois C. Knoll. Graph neural networks for modelling traffic participant interaction. *CoRR*, abs/1903.01254, 2019.
[51] Maximilian Zipfl and J. Marius Zoellner. Towards traffic scene description: The semantic scene graph, 2021.
[52] Matthias Althoff, Markus Koschi, and Stefanie Manzinger. Commonroad: Composable benchmarks for motion planning on roads. In *2017 IEEE Intelligent Vehicles Symposium (IV)*, pages 719–726, 2017.
[53] Volodymyr Mnih and Geoffrey Hinton. Learning to detect roads in high-resolution aerial images. volume 6316, pages 210–223, 12 2010.
[54] Favyen Bastani, Songtao He, Sofiane Abbar, Mohammad Alizadeh, Hari Balakrishnan, Sanjay Chawla, David J. DeWitt, and Sam Madden. Unthule: An incremental graph construction process for robust road map extraction from aerial images. *CoRR*, abs/1802.03680, 2018.
[55] Songtao He, Favyen Bastani, Satvat Jagwani, Mohammad Alizadeh, Hari Balakrishnan, Sanjay Chawla, Mohamed M. Elshrif, Samuel Madden, and Mohammad Amin Sadeghi. Sat2graph: Road graph extraction through graph-tensor encoding. *CoRR*, abs/2007.09547, 2020.
[56] Gellért Máttyus, Wenjie Luo, and Raquel Urtasun. Deeproadmapper: Extracting road topology from aerial images. In *2017 IEEE International Conference on Computer Vision (ICCV)*, pages 3458–3466, 2017.
[57] Yong-Qiang Tan, Shang-Hua Gao, Xuan-Yi Li, Ming-Ming Cheng, and Bo Ren. Vecroad: Point-based iterative graph exploration for road graphs extraction. In *2020 IEEE/CVF Conference on Computer Vision and Pattern Recognition (CVPR)*, pages 8907–8915, 2020.
[58] Min Bai, Gellért Máttyus, Namdar Homayounfar, Shenlong Wang, Shrinidhi Kowshika Lakshmikanth, and Raquel Urtasun. Deep multi-sensor lane detection. *CoRR*, abs/1905.01555, 2019.
[59] Lina Maria Paz, Pedro Piniés, and Paul Newman. A variational approach to online road and path segmentation with monocular vision. In *2015 IEEE International Conference on Robotics and Automation (ICRA)*, pages 1633–1639, 2015.
[60] Namdar Homayounfar, Wei-Chiu Ma, Shrinidhi Kowshika Lakshmikanth, and Raquel Urtasun. Hierarchical recurrent attention networks for structured online maps. In *2018 IEEE/CVF Conference on Computer Vision and Pattern Recognition*, pages 3417–3426, 2018.
[61] Netalee Efrat, Max Bluvstein, Shaul Oron, Dan Levi, Noa Garnett, and Bat El Shlomo. 3d-lanenet+: Anchor free lane detection using a semi-local representation. *CoRR*, abs/2011.01535, 2020.
[62] Namdar Homayounfar, Justin Liang, Wei-Chiu Ma, Jack Fan, Xinyu Wu, and Raquel Urtasun. Dagmapper: Learning to map by discovering lane topology. In *2019 IEEE/CVF International Conference on Computer Vision (ICCV)*, pages 2911–2920, 2019.
[63] Johannes Beck and Christoph Stiller. Non-parametric lane estimation in urban environments. In *2014 IEEE Intelligent Vehicles Symposium Proceedings*, pages 43–48, 2014.
[64] Justin Liang, Namdar Homayounfar, Wei-Chiu Ma, Shenlong Wang, and Raquel Urtasun. Convolutional recurrent network for road boundary extraction. In *2019 IEEE/CVF Conference on Computer Vision and Pattern Recognition (CVPR)*, pages 9504–9513, 2019.
[65] Jannik Zuern, Johan Vertens, and Wolfram Burgard. Lane graph estimation for scene understanding in urban driving, 2021.
[66] Jianwei Yang, Jiasen Lu, Stefan Lee, Dhruv Batra, and Devi Parikh. Graph R-CNN for scene graph generation. *CoRR*, abs/1808.00191, 2018.
[67] J. Douglas Carroll and Phipps Arabie. Multidimensional Scaling. *Annual Review of Psychology*, 31(1):607–649, 1980. 00704 _eprint: https://doi.org/10.1146/annurev.ps.31.020180.003135.
[68] J. Douglas Carroll, Myron Wish, Murray Turoff, and Harold A. Linstone. Vi.c. multidimensional scaling: Models, methods, and relations to delphi. 2002.
[69] James Russell and Merry Bullock. Multidimensional scaling of

emotional facial expressions. similarity from preschoolers to adults. *Journal of Personality and Social Psychology*, 48:1290–1298, 05 1985.

[70] Michael Hornberger, B Bell, K Graham, and Timothy Rogers. Are judgments of semantic relatedness systematically impaired in alzheimer's disease? *Neuropsychologia*, 47:3084–94, 08 2009.

[71] L Pedelty, S C Levine, and S K Shevell. Developmental changes in face processing: results from multidimensional scaling. *J. Exp. Child Psychol.*, 39(3):421–436, June 1985.

[72] Steven Youngentob, Brett Johnson, Michael Leon, Paul Sheehe, and Paul Kent. Predicting odorant quality perceptions from multidimensional scaling of olfactory bulb glomerular activity patterns. *Behavioral neuroscience*, 120:1337–45, 01 2007.

[73] J. Douglas Carroll and Paul E. Green. Psychometric methods in marketing research: Part ii, multidimensional scaling. *Journal of Marketing Research*, 34(2):193–204, 1997.

[74] Norman Kenkel and László Orlóci. Applying metric and nonmetric multidimensional scaling to ecological studies: Some new results. *Ecology*, 67:919–928, 08 1986.

[75] George Rabinowitz. An introduction to nonmetric multidimensional scaling. *American Journal of Political Science*, 19:343, 1975.

[76] Pratik Biswas, Tzu-Chen Lian, Ta-Chung Wang, and Yinyu Ye. Semidefinite programming based algorithms for sensor network localization. *TOSN*, 2:188–220, 05 2006.

[77] Michael C. Hout, Megan H. Papesh, and Stephen D. Goldinger. Multidimensional scaling. *Wiley interdisciplinary reviews. Cognitive science*, 4(1):93–103, January 2013. 00252.

[78] Alexander Bronstein, Michael Bronstein, and Ron Kimmel. Efficient computation of isometry[U+2010]invariant distances between surfaces. *SIAM J. Scientific Computing*, 28:1812–1836, 01 2006.

[79] Amos Tversky. Features of similarity. *Psychological Review*, 84(4):327–352, 1977.

[80] Takayuki Saito and Shin-ichi Takeda. Multidimensional scaling of asymmetric proximity: Model and method. *Behaviormetrika*, 17(28):49–80, Jul 1990.

[81] T Saito. A method of multidimensional scaling to obtain a sphere configuration. *Hokkaido Behavioral Science Report, Series-M*, 4, 1983.

[82] David G Weeks and PM Bentler. Restricted multidimensional scaling models for asymmetric proximities. *Psychometrika*, 47(2):201–208, 1982.

[83] A Okada and T Imaizumi. Geometric models for asymmetric similarity data. *Behaviormetrika*, 21:81–96, 1987.

[84] Akinori Okada and Tadashi Imaizumi. Nonmetric multidimensional scaling of asymmetric proximities. *Behaviormetrika*, 14(21):81–96, 1987.

[85] Gyslain Giguère. Collecting and analyzing data in multidimensional scaling experiments: A guide for psychologists using spss. *Tutorials in Quantitative Methods for Psychology*, 2, 03 2006.

[86] Jan de Leeuw. Modern multidimensional scaling: Theory and applications (second edition). *Journal of Statistical Software*, 14, 10 2005.

[87] Joseph Kruskal and Myron Wish. *Multidimensional Scaling*. SAGE Publications, Inc., 1978.

[88] J. B. Kruskal. Nonmetric multidimensional scaling: A numerical method. *Psychometrika*, 29(2):115–129, June 1964.

[89] John C Gower. Some distance properties of latent root and vector methods used in multivariate analysis. *Biometrika*, 53(3-4):325–338, 1966.

[90] Warren S Torgerson. Multidimensional scaling: I. theory and method. *Psychometrika*, 17(4):401–419, 1952.

[91] Ulrik Brandes and Christian Pich. Eigensolver methods for progressive multidimensional scaling of large data. In *International Symposium on Graph Drawing*, pages 42–53. Springer, 2006.

[92] Ali Civril, Malik Magdon-Ismail, and Eli Bocek-Rivele. Ssde: Fast graph drawing using sampled spectral distance embedding. In *International Symposium on Graph Drawing*, pages 30–41. Springer, 2006.

[93] Vin Silva and Joshua Tenenbaum. Global versus local methods in nonlinear dimensionality reduction. *Advances in neural information processing systems*, 15, 2002.

[94] Jengnan Tzeng, Henry Horng-Shing Lu, and Wen-Hsiung Li. Multidimensional scaling for large genomic data sets. *BMC bioinformatics*, 9(1):1–17, 2008.

[95] Ulrik Brandes and Christian Pich. An experimental study on distance-based graph drawing. In *International Symposium on Graph Drawing*, pages 218–229. Springer, 2008.

[96] Mirza Klimenta and Ulrik Brandes. Graph Drawing by Classical Multidimensional Scaling: New Perspectives. In David Hutchison, Takeo Kanade, Josef Kittler, Jon M. Kleinberg, Friedemann Mattern, John C. Mitchell, Moni Naor, Oscar Nierstrasz, C. Pandu Rangan, Bernhard Steffen, Madhu Sudan, Demetri Terzopoulos, Doug Tygar, Moshe Y. Vardi, Gerhard Weikum, Walter Didimo, and Maurizio Patrignani, editors, *Graph Drawing*, volume 7704, pages 55–66. Springer Berlin Heidelberg, Berlin, Heidelberg, 2013. 00000 Series Title: Lecture Notes in Computer Science.

[97] Andreas Buja, Deborah F Swayne, Michael Littman, Nathaniel Dean, and Heike Hofmann. Xgvis: Interactive data visualization with multidimensional scaling. *Journal of Computational and Graphical Statistics*, pages 1061–8600, 2001.

[98] Jose A. Costa, Neal Patwari, and Alfred O. Hero. Distributed weighted-multidimensional scaling for node localization in sensor networks. *ACM Transactions on Sensor Networks*, 2(1):39–64, February 2006. 00775.

[99] Emden R. Gansner, Yehuda Koren, and Stephen North. Graph Drawing by Stress Majorization. In David Hutchison, Takeo Kanade, Josef Kittler, Jon M. Kleinberg, Friedemann Mattern, John C. Mitchell, Moni Naor, Oscar Nierstrasz, C. Pandu Rangan, Bernhard Steffen, Madhu Sudan, Demetri Terzopoulos, Dough Tygar, Moshe Y. Vardi, Gerhard Weikum, and János Pach, editors, *Graph Drawing*, volume 3383, pages 239–250. Springer Berlin Heidelberg, Berlin, Heidelberg, 2005. 00000 Series Title: Lecture Notes in Computer Science.

[100] Jan De Leeuw. Convergence of the majorization method for multidimensional scaling. *Journal of classification*, 5(2):163–180, 1988.

[101] Ingwer Borg and Patrick JF Groenen. *Modern multidimensional scaling: Theory and applications*. Springer Science & Business Media, 2005.

[102] Jan de Leeuw and Patrick Mair. Multidimensional Scaling Using Majorization: SMACOF in R. *Journal of Statistical Software*, 31(3), 2009. 00000.

[103] Yehuda Koren. Graph drawing by subspace optimization. In *Proceedings of the Sixth Joint Eurographics-IEEE TCVG conference on Visualization*, pages 65–74, 2004.

[104] Joseph B Kruskal. Multidimensional scaling by optimizing goodness of fit to a nonmetric hypothesis. *Psychometrika*, 29(1):1–27, 1964.

[105] Jonathan X. Zheng, Samraat Pawar, and Dan F. M. Goodman. Graph Drawing by Stochastic Gradient Descent. *arXiv:1710.04626 [cs]*, June 2018. arXiv: 1710.04626.

[106] Mark Ortmann, Mirza Klimenta, and Ulrik Brandes. A sparse stress model. In *International Symposium on Graph Drawing and Network Visualization*, pages 18–32. Springer, 2016.

[107] Marc Khoury, Yifan Hu, Shankar Krishnan, and Carlos Scheidegger. Drawing large graphs by low-rank stress majorization. In *Computer Graphics Forum*, volume 31, pages 975–984. Wiley Online Library, 2012.

[108] Timothy A Davis and Yifan Hu. The university of florida sparse matrix collection. *ACM Transactions on Mathematical Software (TOMS)*, 38(1):1–25, 2011.

[109] William H Press, William T Vetterling, Saul A Teukolsky, and Brian P Flannery. *Numerical recipes*, volume 818. Cambridge university press Cambridge, 1986.

[110] Renzo Angles and Claudio Gutierrez. Survey of graph database models. *ACM Computing Surveys*, 40, 02 2008.

[111] William L. Hamilton, Rex Ying, and Jure Leskovec. Representation learning on graphs: Methods and applications. *CoRR*, abs/1709.05584, 2017.

[112] Peter D Hoff, Adrian E Raftery, and Mark S Handcock. Latent space approaches to social network analysis. *Journal of the American Statistical Association*, 97(460):1090–1098, 2002.

[113] Jiyang Gao, Chen Sun, Hang Zhao, Yi Shen, Dragomir Anguelov, Congcong Li, and Cordelia Schmid. Vectornet: Encoding hd maps and agent dynamics from vectorized representation. In *2020 IEEE/CVF Conference on Computer Vision and Pattern Recognition (CVPR)*, pages 11522–11530, 2020.

[114] Ming Liang, Bin Yang, Rui Hu, Yun Chen, Renjie Liao, Song Feng, and Raquel Urtasun. Learning lane graph representations for motion forecasting. *CoRR*, abs/2007.13732, 2020.

[115] Amr Ahmed, Nino Shervashidze, Shravan Narayanamurthy, Vanja Josifovski, and Alexander Smola. Distributed large-scale natural graph factorization. pages 37–48, 05 2013.

[116] Shaosheng Cao, Wei Lu, and Qiongkai Xu. Grarep: Learning graph representations with global structural information. In *Proceedings of the 24th ACM International on Conference on Information and Knowledge Management*, CIKM '15, page 891–900, New York, NY, USA, 2015. Association for Computing Machinery.

[117] Mikhail Belkin and Partha Niyogi. Laplacian eigenmaps and spectral techniques for embedding and clustering. In *NIPS*, 2001.

[118] Bryan Perozzi, Rami Al-Rfou, and Steven Skiena. Deepwalk. *Proceedings of the 20th ACM SIGKDD international conference on Knowledge discovery and data mining*, Aug 2014.

[119] Aditya Grover and Jure Leskovec. node2vec: Scalable feature learning for networks, 2016.

[120] William L. Hamilton, Rex Ying, and Jure Leskovec. Inductive representation learning on large graphs. *CoRR*, abs/1706.02216, 2017.

[121] G. E. Hinton and R. R. Salakhutdinov. Reducing the dimensionality of data with neural networks. *Science*, 313(5786):504–507, 2006.

[122] Shaosheng Cao. deep neural network for learning graph representations. 02 2016.

[123] Daixin Wang, Peng Cui, and Wenwu Zhu. Structural deep network embedding. In *Proceedings of the 22nd ACM SIGKDD International Conference on Knowledge Discovery and Data Mining*, KDD '16, page 1225–1234, New York, NY, USA, 2016. Association for Computing Machinery.

[124] Thomas N. Kipf and Max Welling. Variational graph auto-encoders, 2016.

[125] Thomas N. Kipf and Max Welling. Semi-supervised classification with graph convolutional networks. *CoRR*, abs/1609.02907, 2016.

[126] S. Ingram, T. Munzner, and M. Olano. Glimmer: Multilevel MDS on the GPU. *IEEE Transactions on Visualization and Computer Graphics*, 15(2):249–261, March 2009. 00205.

[127] Nasir Saeed, Haewoon Nam, Tareq Y. Al-Naffouri, and Mohamed-Slim Alouini. A State-of-the-Art Survey on Multidimensional Scaling Based Localization Techniques. *arXiv:1906.03585 [eess]*, June 2019. 00051 arXiv: 1906.03585.

[128] Yi Zhao, Satish V. Ukkusuri, and Jian Lu. Multidimensional Scaling-Based Data Dimension Reduction Method for Application in Short-Term Traffic Flow Prediction for Urban Road Network. *Journal of Advanced Transportation*, 2018:1–10, November 2018. 00005.

[129] Octavian-Eugen Ganea and Gary Bécigneul. Riemannian adaptive optimization methods. In *7th International Conference on Learning Representations (ICLR 2019)*, 2018.

[130] Max Kochurov. K-stereographic projection model., 2018.

[131] Jean-Claude Midler. Non-Euclidean Geographic Spaces: Mapping Functional Distances. *Geographical Analysis*, 14(3):189–203, 1982. 00088 _eprint: https://onlinelibrary.wiley.com/doi/pdf/10.1111/j.1538-4632.1982.tb00068.x.